\documentclass[runningheads]{llncs}

\usepackage[T1]{fontenc}
\usepackage{amsmath}
\usepackage{booktabs}
\usepackage{bbding}
\begin{document}

\title{Scenario Generalization and Sector Transfer in Real Surface Irrigation Surrogate Modelling with Tabular Models}
\titlerunning{Surface Irrigation Surrogate Modelling with Tabular Models}

\author{Alessandro Cagiano\inst{1}\orcidID{0009-0008-2686-1745}\Envelope \and M. Aybars Seyhan\inst{3} \and Fabiola Gangi\inst{2,3} \and Matteo Zignani\inst{1}\orcidID{0000-0002-4808-4106} \and Daniele Masseroni\inst{2,3}}
\authorrunning{A. Cagiano et al.}

\institute{
Department of Computer Science, University of Milan, Milan, Italy\\
\email{\{alessandro.cagiano,matteo.zignani\}@unimi.it}
\and
Department of Agricultural and Environmental Sciences, University of Milan, Milan, Italy
\and
Digital Drop s.r.l., Milan, Italy
}
\maketitle

\begin{abstract}
Surface irrigation planning requires screening many operating conditions and spatial configurations, while high-fidelity shallow-water-equation (SWE) simulations are expensive. We study tabular surrogate models for predicting maximum water height from drone-derived field morphology, terrain, hydrological, spatial, and scenario descriptors, where each row represents a spatial cell under one irrigation scenario. To assess deployment feasibility, we evaluate gradient-boosted trees, neural tabular models, AutoML, and tabular foundation models (TFMs) under two regimes: same-sector generalization, where a model predicts unseen scenarios after observing a limited set of configurations in the same sector, and sector-out transfer, where knowledge learned from source sectors is transferred to an unseen sector. The main findings are twofold: \textit{a}) under the evaluated fixed 8/32 same-sector split, surrogates achieve strong performance, especially with raw spatial coordinates; \textit{b)} sector-out transfer is substantially harder because the target sector has different coordinates and geometry. Sector-frame normalization of spatial coordinates improves transfer across similarly structured sectors, but does not solve cross-shape generalization. Overall, tabular surrogates appear useful for same-sector scenario screening under this fixed split, while robust transfer to unseen sectors remains open.
\keywords{Tabular foundation models \and Surface irrigation surrogate modelling \and Domain shift \and Spatial transfer \and Scenario generalization}
\end{abstract}

\section{Introduction}\label{sec:intro}

Surface irrigation planning requires evaluating how water propagates over field sectors under different operating conditions, such as irrigation duration and flow rate. High-fidelity shallow-water-equation simulations provide detailed estimates of this behaviour, but they are costly when many scenarios must be screened for the same terrain. At the same time, drone-derived field morphology, terrain descriptors, hydrological variables, and scenario-specific settings make it possible to represent each sector as a large tabular prediction problem, where each row corresponds to one spatial cell under one irrigation scenario and the target is the maximum simulated water height at that cell.

The key methodological difficulty is that this is not an ordinary independent and identically distributed tabular setting. Rows are spatially repeated, cells are observed under multiple interventions, and scenarios share the same underlying sector geometry. Random row splits would therefore mix nearby locations and related scenario structures across train and test sets, providing an optimistic view of deployment performance. A meaningful evaluation must instead distinguish between predicting unseen scenarios on an already observed sector and transferring knowledge to a sector whose geometry has not been observed during training.

These two settings correspond to different deployment questions. In same-sector generalization, a limited number of expensive SWE simulations are available for a target sector, and the surrogate is used to estimate the remaining operating scenarios on the same geometry. Such a setting is relevant during the deployment of software platforms that assist decision makers and farmers in deciding the best surface irrigation strategy in a quasi-real-time context. In this setting, raw spatial coordinates are legitimate features because they identify stable locations within the observed domain. In sector-out transfer, by contrast, the target sector is held out during training. Raw absolute coordinates are then unsafe as primary predictors, since the same numerical coordinate may correspond to unrelated hydrological roles in different sectors. Transfer, therefore, requires feature representations whose meaning is more portable across domains.

In this paper, we evaluate tabular surrogate models on real irrigation-sector data under these two deployment regimes. The benchmark includes three aligned sectors with a common 40-scenario grid, used for scenario-disjoint same-sector evaluation and aligned sector-out transfer, and an additional cross-shape domain with different geometry and scenario configuration, used as a stress test for spatial generalization. We compare gradient-boosted trees, neural tabular models, tabular foundation models, and an AutoML ensemble control under explicit sampling and evaluation constraints. Because some high-capacity and foundation-model evaluations are constrained by memory or inference cost, we distinguish full-table results from primary-sample evaluations whenever needed. We also study coordinate handling through no-raw-coordinate transfer, per-sector min–max scaling, and sector-frame normalization, and audit whether engineered spatial descriptors such as proximity from the irrigation gate remain portable across sectors.

The results show a clear separation between the two deployment regimes. Same-sector surrogates achieve high accuracy when the target geometry is observed, especially with raw coordinates, and can provide substantially faster inference than full SWE simulation for scenario screening. Sector-out transfer is substantially harder: removing raw coordinates avoids a non-portable shortcut, while sector-frame normalization improves transfer across similarly structured aligned sectors but does not solve cross-shape generalization. Model rankings are also not stable across target sectors, indicating that current tabular models do not yet provide a universally robust solution to unseen-sector deployment.

Overall, the benchmark supports tabular surrogate modelling as a practical near-term tool for reducing the number of SWE simulations required within an observed irrigation sector. At the same time, it identifies robust spatial transfer as the main open challenge. Progress in this direction will likely require shape-aware spatial descriptors, physically grounded representations of boundary and flow structure, uncertainty estimation, and active scenario selection strategies for deciding which expensive simulations should be acquired.

\section{Related Work}\label{sec:related}

Gradient-boosted decision trees remain strong tabular baselines: XGBoost~\cite{chen2016xgboost}, LightGBM~\cite{ke2017lightgbm}, and CatBoost~\cite{prokhorenkova2018catboost} combine accuracy with practical training and inference behaviour, while AutoGluon-Tabular~\cite{erickson2020autogluon} searches and ensembles across model families. Neural tabular models have also improved: RealMLP provides strong pre-tuned MLP defaults~\cite{holzmueller2024better}, TabM uses parameter-efficient ensembling~\cite{gorishniy2024tabm}, and TFMs such as TabPFN variants~\cite{hollmann2025tabpfn,grinsztajn2025tabpfn25,grinsztajn2026tabpfn3} and TabICLv2~\cite{qu2026tabiclv2} reuse learned priors or context-based prediction mechanisms. These methods are relevant because irrigation surrogate data are large, mostly numerical, and contain nonlinear interactions among terrain, flow, and scenario variables. General tabular rankings, however, do not directly answer the deployment question studied here: irrigation-sector data are spatially repeated, scenario-indexed, and evaluated under sector shift rather than random IID row splits. A model may rank highly on public tabular datasets while exploiting non-portable location features, so we compare model families under shared scenario-disjoint and sector-out conditions.

The application connects to surrogate modelling for expensive physical or hydrological simulations~\cite{dai2024hydrologic}, including SWE surrogates for hydraulic flow fields~\cite{song2023swe}. Sector-out transfer is also closer to domain generalization than to ordinary IID prediction: the task is fixed, but the target spatial domain is unseen during training~\cite{gulrajani2021domainbed}. Unlike many remote-sensing transfer benchmarks, the target is a continuous physical response induced by an irrigation scenario; the same spatial cell appears under multiple interventions, so a useful benchmark must preserve both spatial and scenario structure.

\section{Data Sources and Task}\label{sec:data}

To train and test the different surrogate models, we combine two sources. First, drone imagery is preprocessed to obtain a tabular representation of each sector: the field morphology is discretized into spatial cells, and each cell is described by terrain, hydrological, and spatial descriptors. Second, a two-dimensional shallow-water-equation (SWE) simulator evaluates irrigation scenarios over the same sector and provides, for each cell, the maximum water height reached during the scenario. This simulated maximum height is the target variable that the surrogate model must estimate with substantially lower inference time than the full physical simulation, which requires approximately two hours per scenario and sector in the workflow. The fundamental spatial unit in this study is an \textit{irrigation sector}, defined as a small portion of a field with its own geometry and morphology.  

In the resulting tabular representation, each row describes one spatial cell under one irrigation scenario. The regression target is encoded by the \texttt{target\_maxh} column, which corresponds to the simulated maximum water height at that cell. Each scenario corresponds to a specific irrigation configuration defined by different a) irrigation duration and b) flow-rate settings. Cell features include terrain and hydrological descriptors (\texttt{Slope}, \texttt{FlowAcc}, \texttt{FlowDir}, \texttt{Aspect}), scenario descriptors such as irrigation duration and flow rate, angular encodings of flow direction and aspect, proximity descriptors, and, in coordinate-inclusive protocols (see Section \ref{sec:protocols}), raw \texttt{x}, \texttt{y}, and \texttt{z}.

In this study, we focus on three real irrigation sectors, denoted as aligned sectors A, B, and C, each equipped with 40 simulated scenarios. We also include a fourth real cross-shape domain with triangular geometry. This sector has a different shape and scenario configurations, so it is used as a cross-shape stress test where appropriate. Tables for sectors A, B and C contain 16.1M, 18.9M, and 17.1M rows, respectively; while the cross shape domain contains 4.0M rows over eight scenarios. The cross-shape sector is treated differently because its scenario configuration and geometry do not support the same 40-scenario sector protocol. This distinction prevents a visually appealing aggregate table from mixing non-equivalent experimental questions.

Feature sets are tied to the deployment question and change across the different evaluation settings. For instance, raw coordinates are allowed for same-sector prediction because the model is predicting new scenarios over an observed geometry. For primary transfer, raw coordinates are removed. This does not remove all spatial information: terrain-derived features such as slope, aspect, flow accumulation, and proximity remain. The phrase ``no raw XYZ'' therefore means that absolute coordinates are excluded, not that the representation is non-spatial.

\section{Evaluation Protocol}\label{sec:protocols}

The feasibility of surrogate models as a replacement for SWE simulations to inform surface irrigation strategies in a deployment setting passes through evaluation protocols that measure whether tabular surrogates can deal with two critical points. The first is \textit{scenario reduction}: given a sector for which a limited number of SWE evaluations are available, the model should estimate the response of additional operating scenarios over the same field/sector geometry. The importance of the point lies in the possibility of considerably reducing the number of scenarios to be simulated and exploiting the surrogate model as an interpolating method to generate/predict the remaining scenarios. From a deployment perspective, it results in a sensible reduction of the setup time and latency when predictions are required for guiding the surface irrigation processes. The second point is \textit{sector model transfer}: the model should reuse relationships learned from source sectors to estimate responses on a held-out sector. It translates into evaluating the generalization capability of the model on unseen and new sectors, which may differ from the training sectors on diverse aspects, such as the field geometry, the field dimensions, or the soil characteristics. 

\noindent\textit{Same-sector generalization.} To deal with the first critical point, for the A, B and C sectors, we trained the tabular models on eight scenarios and tested on the remaining ones (32). The eight training scenarios are a fixed coverage split of the 40-scenario duration/flow-rate grid, reused for all aligned sectors and not resampled across runs. This fixed split supports controlled comparison across model families. To avoid data leakage, the split is scenario-disjoint: no row from a held-out scenario is used during training. Metrics are computed on a deterministic primary sample shared across models and, when feasible, on the full held-out table. In same-sector experiments, the primary sample contains 16k cells per held-out scenario, i.e., 512k cells over the 32 test scenarios. This lets expensive TFM variants be compared on identical rows. Where training subsampling is required, it preserves scenario coverage and target-value variation rather than using arbitrary row prefixes. To ensure comparability with performances obtained by past SWE model outcomes, the main metric is $R^2$, with RMSE used during run inspection but not reported in the following main tables. We report full-table scores when they are available under the final protocol. When a model family is evaluated only on the primary sample, this is marked explicitly.

\noindent\textit{Sector-out transfer.} The second critical point defines an evaluation protocol to assess the generalization capability on unseen and new sector data. In this setting, a target sector is held out and the model is trained on the remaining aligned sectors. The primary transfer metric is computed on the same deterministic held-out cell sample for every model: 16k cells per scenario, i.e., 640k cells over the 40 held-out-sector scenarios. When building the training set, we applied source sampling by stratifying on source domain, scenario, and target-value bins, with explicit handling of zero and near-zero targets. In the generation of the training set, we impose a budget on the training set size since source tables are large enough that not every model can consume all rows. By leveraging the above sampling strategy even models with capacity limits receive stratified subsamples rather than arbitrary prefixes or random samples dominated by frequent target values. This defines a best-feasible configuration under model constraints while keeping the sampling policy explicit. Finally, we exclude raw \texttt{x}, \texttt{y}, and \texttt{z} from the cell feature set, but we evaluate three different options for handling coordinates. The first is the primary no-raw-XYZ transfer feature set, where we drop the coordinate information. The second adds per-sector min--max transformation on coordinates, which controls the numerical range of each axis. The third uses sector-frame normalized coordinates:
\begin{align}
x^{\mathrm{sf}}_{i,s}
  &= \frac{x_{i,s} - \frac{1}{2}(x^{\min}_{s}+x^{\max}_{s})}{d_s},
&
y^{\mathrm{sf}}_{i,s}
  &= \frac{y_{i,s} - \frac{1}{2}(y^{\min}_{s}+y^{\max}_{s})}{d_s},
\end{align}
\begin{align}
z^{\mathrm{sf}}_{i,s}
  &= \frac{
      z_{i,s} - \operatorname{median}_{j\in s}(z_{j,s})
    }{
      \operatorname{IQR}_{j\in s}(z_{j,s})+\epsilon
    },
&
d_s
  &= \sqrt{
      (x^{\max}_{s}-x^{\min}_{s})^2
      +
      (y^{\max}_{s}-y^{\min}_{s})^2
    }+\epsilon .
\end{align}

Here $x_s^{\min},x_s^{\max},y_s^{\min},y_s^{\max}$ are sector-wise coordinate extrema over cells, $j\in s$ indexes cells in sector $s$, $\mathrm{sf}$ denotes sector-frame coordinates, and $\epsilon$ is a small constant used for numerical stability. The planar coordinates \texttt{x} and \texttt{y} are centered by the sector bounding-box midpoint and scaled by the bounding-box diagonal, while elevation (\texttt{z}) is standardized by the sector median and inter-quartile range. The resulting coordinates are therefore expressed relative to each sector's own spatial extent. This transformation uses target-domain geometry but no target labels, so it should be interpreted as an unlabeled-domain alignment diagnostic. It tests whether relative position in similarly structured sectors is useful, not whether the representation is physically shape-invariant.

Since the engineered proximity descriptor is spatial, it may encode useful terrain context or a non-portable domain shortcut. We therefore repeat the sector-frame transfer protocol after removing only this descriptor. To guarantee the comparability across experiments, model training and runs, folds, seeds, target, training budget, and primary evaluation rows are kept fixed.

\section{Models}

In this study, we evaluate a set of models suitable for data represented as a table. We group these models into three groups according to the model type, i.e., gradient-boosted or neural methods, and the role, i.e., baseline and standard or foundation models. The gradient-boosted baselines are XGBoost, LightGBM, and CatBoost, while the neural tabular baselines are RealMLP-TD, TabM-D, and TorchMLP. The tabular foundation models - TFMs - group includes RealTabPFN-2.5, TabPFN-2.6, TabPFN-3, TabPFN-3-OOD, and TabICLv2. TabPFN entries are separated because releases differ in weights, preprocessing, and inference. According to a well-established protocol in the TFM literature, we also include AutoGluon 1.5 Extreme as a primary-only AutoML ensemble baseline on the best transfer feature configuration. These families are compared under the same splits with a best-feasible training budget: models with smaller context or memory capacity receive stratified samples. This grouping is useful because the families make different promises and are based on different paradigms. Gradient-boosted models provide strong practical baselines. Neural tabular models assess whether learned representations outperform gradient-boosted methods when many sector rows are available. TFM models test whether a pretrained tabular prior or context mechanism transfers to this scientific domain. Finally, using AutoGluon, we test whether an automated ensemble of modern tabular components changes the conclusions reached by individual model families.

\section{Results}

\noindent\textbf{Same-sector generalization. }
Table~\ref{tab:same-sector-summary} summarizes same-sector generalization: given simulations for eight scenarios, we predict unseen scenarios in the same sector. With coordinates, all three aligned sectors reach high full-table accuracy: $R^2=0.9830$ on sector A, $R^2=0.9705$ on sector B, and $R^2=0.9571$ on sector C. Neural tabular models are the strongest completed full-table family, although gradient-boosted baselines and TFMs remain competitive. Scenario-level diagnostics also show stronger robustness for neural and TFM models: for each family, the highest worst-case $R^2$ across the thirty-two held-out scenarios is 0.758/0.934/0.940 on sector A, 0.762/0.931/0.867 on sector B, and 0.707/0.885/0.857 on sector C for gradient-boosted, neural models and TFMs, respectively.
\begin{table}[h!]
\centering
\caption{Same-sector prediction summary. Values are $R^2$; $\dagger$ marks
primary-sample results and $\ddagger$ marks a same-scenario held-out-cell split.}
\label{tab:same-sector-summary}
\begin{tabular}{llrrr}
\toprule
Domain & Feature set & Best gradient-boosted & Best neural & Best TFM \\
\midrule
Aligned sector A & raw XYZ & 0.9461 & \textbf{0.9830} & 0.9642 \\
Aligned sector B & raw XYZ & 0.8892 & \textbf{0.9705} & 0.9406$^\dagger$ \\
Aligned sector C & raw XYZ & 0.8769 & \textbf{0.9571} & 0.9319$^\dagger$ \\
Cross-shape domain$^\ddagger$ & raw XYZ, held-out cells & 0.9965 & 0.9965 & \textbf{0.9978} \\
Aligned sector A & no raw XYZ & 0.7615 & \textbf{0.7699} & 0.7162 \\
\bottomrule
\end{tabular}
\vspace{-0.15in}
\end{table}

Finally, the last row in Table~\ref{tab:same-sector-summary}, no-raw-XYZ, shows that removing raw coordinates reduces the best full-table $R^2$ on sector A from 0.9830 to 0.7699. This does not make raw coordinates invalid: in the same-sector prediction, they index stable local terrain patterns, whereas in sector transfer, they can become non-portable features. The drop also indicates that high same-sector accuracy partly relies on stable local spatial identity, rather than on a coordinate-free hydrological rule. Because the same-sector use case is intended to reduce the number of expensive SWE evaluations, inference time is part of the practical comparison. In Table~\ref{tab:inference-times}, we report the corresponding same-sector wall-clock timings for sector A. In general, we notice a significant improvement in the inference time for most of the models with respect to the simulation time of the SWE. By combining the above findings, we can assess surrogate models as a valid and more efficient replacement for same-sector generalization.
\begin{table}[t]
\centering
\caption{Inference time on aligned sector A. Full-table evaluation uses
$\sim$12.7M held-out cells; the primary evaluation uses 512k cells. LightGBM
uses GPU training, but prediction is CPU-backed.}
\label{tab:inference-times}
\small
\begin{tabular}{llrr}
\toprule
Model & Eval. set & Inference time & $R^2$ \\
\midrule
XGBoost & full table & 3.6 s & 0.9354 \\
LightGBM & full table & 99.9 s & 0.9461 \\
CatBoost & full table & \textbf{3.1 s} & 0.9102 \\
RealMLP-TD & full table & 12.2 s & 0.9751 \\
TabM-D & full table & 61.2 s & \textbf{0.9830} \\
RealTabPFN-2.5 & full table & 7203.2 s & 0.9642 \\
TabPFN-2.6 & full table & 17337.7 s & 0.9606 \\
TabPFN-3 & 512k primary & 2018.9 s & 0.9472 \\
TabICLv2 & 512k primary & 1061.2 s & 0.9416 \\
\bottomrule
\end{tabular}
\vspace{-0.05in}
\end{table}

\noindent\textbf{Sector-out transfer. } As for the second critical point, i.e., the generalization capability on unseen sectors, Table~\ref{tab:sector-transfer-no-xyz} reports the primary transfer protocol where we drop raw XYZ coordinates. Scores are much lower than in the previous setting, with best held-out $R^2$ values of 0.5885 for aligned sector A, 0.4625 for aligned sector B, and 0.6836 for aligned sector C. The model category also changes by target sector: a TFM variant is best on aligned sectors A and C, while LightGBM is best on aligned sector B. This instability is important because it argues against a simple claim that one model class solves sector transfer: the three best models are TabPFN-2.6, LightGBM, and TabPFN-3-OOD.

The absolute scale of the transfer scores is also informative. Even the best model for no-raw-XYZ transfer remains far below the same-sector setting where the target geometry is observed. The models learn useful terrain and scenario relationships, but these relationships are not sufficient to recover the target-sector response with the same accuracy as an observed-sector surrogate.
\begin{table}[t]
\centering
\caption{Sector-out transfer with raw coordinates excluded. Each value is
primary-sample $R^2$ on the held-out aligned sector. Source training uses the
remaining aligned sectors with a stratified source-sample budget.}
\label{tab:sector-transfer-no-xyz}
\begin{tabular}{lrrrr}
\toprule
Held-out sector & Best gradient-boosted & Best neural & Best TFM & Best overall \\
\midrule
Aligned sector A & 0.5851 & 0.5639 & \textbf{0.5885} & 0.5885 \\
Aligned sector B & \textbf{0.4625} & 0.4552 & 0.4375 & 0.4625 \\
Aligned sector C & 0.6468 & 0.6093 & \textbf{0.6836} & 0.6836 \\
\bottomrule
\end{tabular}
\vspace{-0.15in}
\end{table}

Therefore, sector-out transfer is not merely a lower-accuracy version of the same-sector prediction; it is a qualitatively different generalization problem that, given the overall poor performance, still represents an open issue.

\noindent\textbf{Coordinate transformations. } We also investigate which is the most effective transformation to handle the spatial coordinates of the cells. Table~\ref{tab:coordinate-diagnostics} isolates coordinate handling using gradient-boosted baselines. We observe that the per-sector min--max transformation is not sufficient for transferring and reducing the mean $R^2$ relative to the no-raw-XYZ baseline. On the contrary, sector-frame normalization substantially improves aligned-sector transfer, increasing the mean $R^2$ from 0.5319 to 0.7526 for the gradient-boosted models. This indicates that relative position can be useful when sectors are similarly structured.

The same sector-frame representation does not extend to the cross-shape domain, i.e., when applied to the fourth sector. In this setting, the mean $R^2$ values for gradient-boosted methods are -0.4966 without raw coordinates, -3.6958 with per-sector min--max coordinates, and -5.1340 with sector-frame normalized coordinates. This suggests that the cross-shape gap is not simply a matter of coordinate scale; when the target shape changes, a normalized location can have a different hydrological meaning. Because the cross-shape domain differs both in geometry and in scenario configuration, this result should be read as a stress test rather than a controlled shape-only ablation. Shape-aware descriptors such as boundary distance, source or outlet distance, along-flow distance, or physically grounded corner/edge distances are natural future directions.
\begin{table}[t]
\centering
\caption{Aligned-sector coordinate diagnostic using gradient-boosted baselines. Values are
mean primary-sample $R^2$ across XGBoost, LightGBM, and CatBoost with a
matched 300k target-stratified training budget. Delta is measured relative to
the no-raw-XYZ transfer baseline.}
\label{tab:coordinate-diagnostics}
\begin{tabular}{lrr}
\toprule
Coordinate protocol & Mean $R^2$ & $\Delta$\\
\midrule
No raw XYZ & 0.5319 & -- \\
Per-sector min--max XYZ & -0.7010 & -1.2329 \\
Sector-frame normalized XYZ & \textbf{0.7526} & +0.2207  \\
\bottomrule
\end{tabular}
\vspace{-0.05in}
\end{table}

\noindent\textbf{Sector-frame neural and TFM transfer. } Table~\ref{tab:sector-frame-neural-tfm} reports transfer results with sector-frame normalized coordinates for neural tabular and TFM models. TabPFN-2.6 has the best mean primary $R^2$ among these families, but the ranking is fold-dependent: TabPFN-3 is strongest on aligned sector B, while TorchMLP and RealMLP are competitive on individual folds. This supports the broader interpretation that sector-relative coordinates make aligned transfer more viable without making one model family uniformly dominant.
\begin{table}[t]
\centering
\caption{Sector-frame transfer results for neural tabular and TFM models.
Values are primary-sample $R^2$ on the held-out aligned sector; the mean is
computed over the three aligned sector-out folds.}
\label{tab:sector-frame-neural-tfm}
\small
\begin{tabular}{llrrrr}
\toprule
Family & Model & Aligned A & Aligned B & Aligned C & Mean \\
\midrule
Neural & RealMLP-TD & 0.5312 & 0.8196 & 0.7419 & 0.6976 \\
Neural & TabM-D & 0.5730 & 0.4143 & 0.7824 & 0.5899 \\
Neural & TorchMLP & 0.6206 & 0.8011 & 0.7699 & 0.7305 \\
TFM & RealTabPFN-2.5 & 0.5336 & 0.7733 & 0.7993 & 0.7020 \\
TFM & TabICLv2 & 0.5666 & 0.7254 & 0.6893 & 0.6604 \\
TFM & TabPFN-2.6 & \textbf{0.6651} & 0.8025 & \textbf{0.8257} & \textbf{0.7644} \\
TFM & TabPFN-3 & 0.5058 & \textbf{0.8340} & 0.7408 & 0.6935 \\
TFM & TabPFN-3-OOD & 0.5333 & 0.7769 & 0.6919 & 0.6674 \\
\bottomrule
\end{tabular}
\vspace{-0.15in}
\end{table}

Finally, we investigate the role of the proximity feature. In fact, a proximity-descriptor ablation improves mean primary $R^2$ for all model groups, suggesting that spatial descriptors should be audited for portability. AutoGluon Extreme reaches a mean primary $R^2=0.7676$ on the same folds, supporting the findings related to the sector-out transfer and cross-shape.

\section{Discussion and Conclusions}

The experiments highlight a clear separation between the two deployment regimes. Same-sector surrogate modelling appears effective under the evaluated fixed split when the target geometry is observed: simulations from eight scenarios support accurate prediction of the remaining thirty-two, particularly when raw coordinates are included as stable within-sector location identifiers. By contrast, sector-out transfer remains substantially more difficult. Removing raw coordinates avoids a non-portable shortcut, while sector-frame normalization improves transfer across similarly structured aligned sectors. However, the cross-shape domain shows that relative coordinates alone do not provide shape-invariant spatial representations. The main limitations concern benchmark breadth, split variability, and computational scale. The aligned evaluation includes three sectors sharing the same 40-scenario grid, and the same-sector results are based on one fixed 8/32 scenario split rather than repeated scenario selections. Some TFM results are necessarily restricted to primary-sample evaluation. Broader validation should include additional geometries, soil conditions, hydraulic regimes, wet-region error diagnostics, and active scenario selection to identify the most informative SWE simulations.

Overall, this paper provides a real-data benchmark for irrigation-sector surrogate modelling with tabular models. The results support same-sector scenario generalization as a practical near-term use case for reducing simulation cost, while robust unseen-sector and cross-shape deployment remains an open challenge requiring stronger shape-aware representations and broader validation.

\bibliographystyle{splncs04}
\bibliography{references}

\end{document}